\documentclass[]{spie}  

\usepackage{amsmath,amsfonts,amssymb}
\usepackage{subcaption}
\usepackage{graphicx}
\usepackage{cite} 
\usepackage{epsfig}
\usepackage{amsmath}
\usepackage{nccmath}
\usepackage{amssymb}
\usepackage{mwe}
\usepackage{acro}
\usepackage{amssymb}
\usepackage{xcolor,colortbl}
\usepackage{tabularx}
\usepackage{relsize}
\usepackage{pifont}
\usepackage{booktabs} 
\usepackage{multirow}
\usepackage{multicol}
\usepackage{adjustbox}
\usepackage{float}
\usepackage{graphicx}
\usepackage{makecell}
\usepackage{tabu}
\usepackage[colorlinks=true, allcolors=blue]{hyperref}
\usepackage[capitalize]{cleveref}
\usepackage{tabularray} 
\usepackage{booktabs} 

\title{Img2ST-Net: Efficient High-Resolution Spatial Omics Prediction from Whole Slide Histology Images via Fully Convolutional Image-to-Image Learning}

\author[a]{Junchao Zhu}
\author[c]{Ruining Deng}
\author[b]{Junlin Guo}
\author[a]{Tianyuan Yao}
\author[b]{Juming Xiong}
\author[b]{Chongyu Qu}
\author[e]{Mengmeng Yin}
\author[e]{Yu Wang}
\author[e]{Shilin Zhao}
\author[e]{Haichun Yang}
\author[f]{Daguang Xu}
\author[f]{Yucheng Tang *}
\author[a,b]{Yuankai Huo *}

\affil[a]{Department of Computer Science, Vanderbilt University, Nashville, TN, USA}
\affil[b]{Department of Electrical and Computer Engineering, Vanderbilt University, Nashville, TN, USA}
\affil[c]{ Weill Cornell Medicine, NY, USA}
\affil[d]{Department of Pathology, Microbiology and Immunology, Vanderbilt University Medical Center, Nashville, TN, USA}
\affil[e]{Department of Biostatistics, Vanderbilt University Medical Center, Nashville, TN, USA}
\affil[f]{NVIDIA, CA, USA}

\authorinfo{Corresponding author: Yuankai Huo: E-mail: yuankai.huo@vanderbilt.edu}

\begin{document} 
\maketitle
\begin{abstract}
Recent advances in multi-modal AI have demonstrated promising potential for generating the currently expensive spatial transcriptomics (ST) data directly from routine histology images, offering a means to reduce the high cost and time-intensive nature of ST data acquisition. However, the increasing resolution of ST—particularly with platforms such as Visium HD achieving 8~$\mu$m or finer—introduces significant computational and modeling challenges. Conventional spot-by-spot sequential regression frameworks become inefficient and unstable at this scale, while the inherent extreme sparsity and low expression levels of high-resolution ST further complicate both prediction and evaluation. To address these limitations, we propose Img2ST-Net, a novel high-definition (HD) histology-to-ST generation framework for efficient and parallel high-resolution spatial transcriptomics (ST) prediction. Unlike conventional spot-by-spot inference methods, Img2ST-Net employs a fully convolutional architecture to generate dense, HD gene expression maps in a parallelized manner. By modeling HD ST data as super-pixel representations, the task is reformulated from image-to-omics inference into a super-content image generation problem with hundreds or thousands of output channels. This design not only improves computational efficiency but also better preserves the spatial organization intrinsic to spatial omics data. To enhance robustness under sparse expression patterns, we further introduce SSIM-ST, a structural-similarity-based evaluation metric tailored for high-resolution ST analysis. We present a scalable, biologically coherent framework for high-resolution spatial transcriptomics prediction. Img2ST-Net offers a principled solution for efficient and accurate ST inference at scale. Our contributions lay the groundwork for next-generation ST modeling that is robust and resolution-aware.  The source code has been made publicly available at \url{https://github.com/hrlblab/Img2ST-Net}.

\end{abstract}


\section{Introduction}
The emergence of spatial transcriptomics (ST) has introduced a powerful paradigm for research on disease mechanisms, enabling the joint analysis of tissue morphology and gene expression patterns~\cite{burgess2019spatial, zhu2025asign}. By capturing spatially resolved transcriptomic profiles across tissue sections, ST provides valuable insights into cell–microenvironment interactions and supports the development of more precise therapeutic strategies~\cite{asp2019spatiotemporal, asp2020spatially}. However, the high cost and technical complexity of ST sequencing limit its scalability and accessibility in large-scale applications. Meanwhile, whole-slide images (WSIs) from pathology, which are routinely available in clinical workflows, have been shown to exhibit strong spatial correlations with gene expression patterns~\cite{badea2020identifying, he2020integrating}. This observation has motivated a growing body of research that leverages deep learning to directly infer spatial gene expression from histopathology images. Such image-based methods offer a low-cost, non-destructive alternative to traditional ST sequencing and open up new possibilities for scalable spatial transcriptomic analysis in real-world settings.

\begin{figure*}
    \centering
\includegraphics[width=\linewidth]{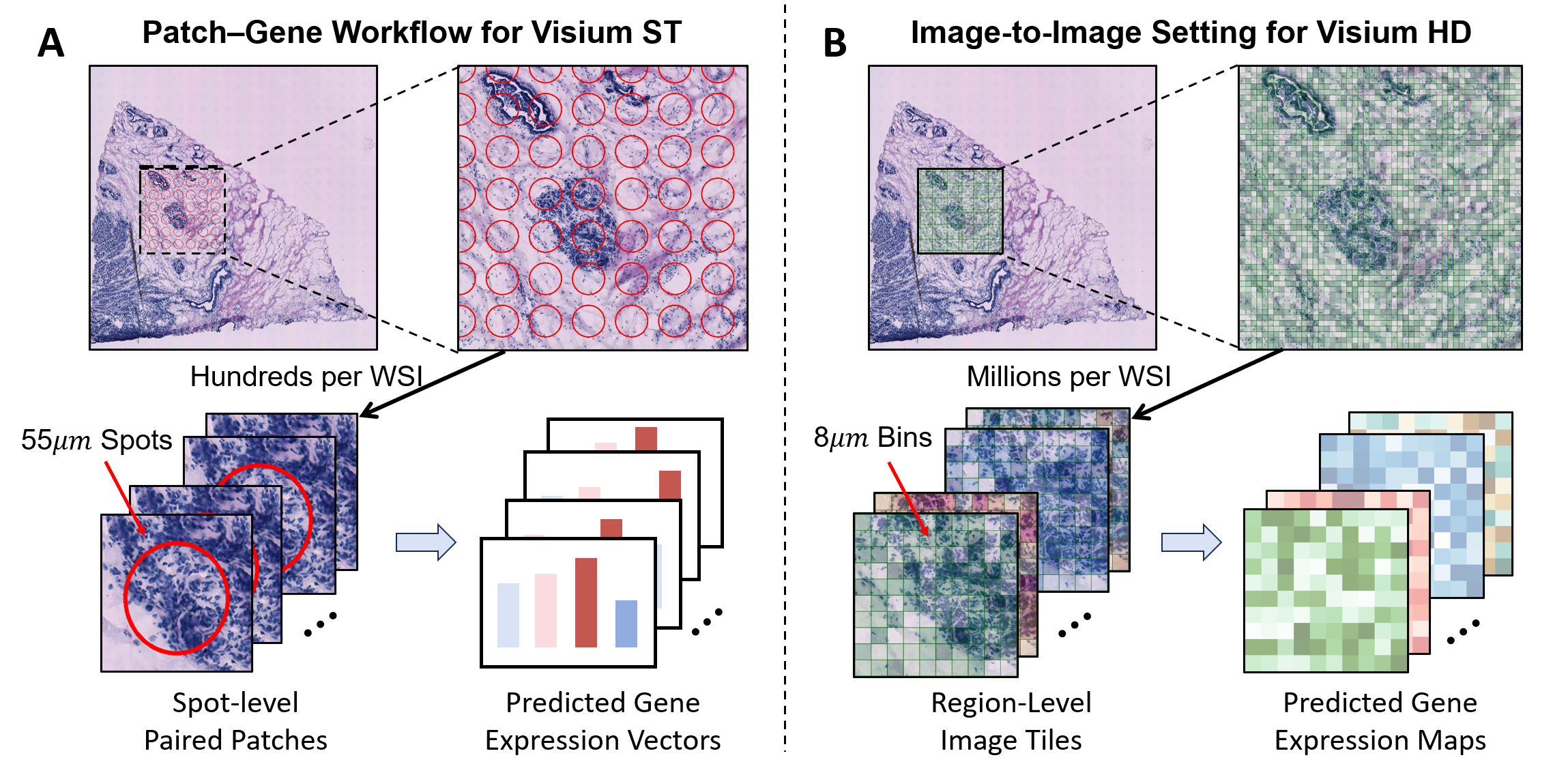}
    \caption{\textbf{Modeling paradigm for spatial transcriptomics prediction.}
 (A) Conventional patch-to-spot regression manner for Visium ST data: each WSI contains hundreds of 55~$\mu$m spots for the ST slide. A separate gene expression vector is predicted for each spot from its corresponding image patch.
 (B) Our proposed image-to-image prediction framework for Visium HD data: each WSI contains millions of 8~$\mu$m  bins for the HD slide. A region-wise modeling strategy where each image region covers multiple bins is used to predict a high-resolution gene expression map, which enables more fine-grained and computationally efficient inference.
}
    \label{fig:Demo}
\end{figure*}

The growing success of deep learning in medical image analysis~\cite{ke2023clusterseg,zhu2023anti,qu2025post,xiong2025zeroreg3d} has spurred the development of methods for ST inference from histological images. Existing approaches adopt a spot-wise prediction strategy, where patches are extracted around individual spot centers and mapped to corresponding gene expression values. By leveraging architectures such as CNNs~\cite{he2020integrating,yang2023exemplar}, Transformers~\cite{}, and graph neural networks (GNNs)~\cite{pang2021leveraging,zeng2022spatial,jia2024thitogene}, these methods consider ST prediction task as a regression task and aim to capture spatial dependencies~\cite{zeng2022spatial,pang2021leveraging} and inter-region visual similarities~\cite{zeng2022spatial,pang2021leveraging,xie2024spatially} through deep feature extraction. However, regression-based models struggle to model the complex and high-dimensional distributions of gene expression data, thus limiting their performance in terms of precision and generalizability. Recent work such as STEM~\cite{zhu2025diffusion} introduces a generative framework based on diffusion models, where pathology images serve as conditions to synthesize spatial expression profiles. 

With the rapid advancement of ST technologies~\cite{staahl2016visualization, wang2018three, eng2019transcriptome}, the spatial resolution of ST data has improved dramatically. Early ST data offered spot-level resolution at 55$\mu$m, while recent developments such as Visium HD now achieve ultra-high resolutions of 8$\mu$m or even 2$\mu$m, which nears single-cell resolution~\cite{benjamin2024multiscale, oliveira2024characterization, janesick2023high}. However, resolution boost introduces substantial computational challenges. For each WSI, which may contain tens of thousands of pixels, the number of gene expression sampling points has increased from a few hundred in ST data to even millions of HD bins. Existing methods that rely on pair-wise mapping paradigms—associating each expression site with a corresponding image region—become infeasible at this scale. Such approaches incur high computational and memory costs and result in unstable training or degraded performance. As ST technologies continue to evolve, there is an urgent need for efficient strategies that can accommodate the growing data scale while maintaining accuracy and robustness.

Besides, current Visium HD platforms offer spatial granularity comparable to that of single-cell RNA sequencing. Such advancement introduces a pronounced sparsity in gene expression patterns, mirroring the “sparse expression” problem widely reported in scRNA-seq datasets~\cite{asp2019spatiotemporal}. As the scale of the data grows and expression values become increasingly sparse and low in magnitude, traditional evaluation metrics such as the Pearson correlation coefficient (PCC) face serious limitations in reliably assessing model performance. Specifically, when most ground-truth values are near zero, even minor fluctuations or noise in the predictions can cause large deviations in PCC, reducing its numerical stability and rendering it insensitive to meaningful differences in model output. These challenges underscore the need for more robust evaluation metrics that can better capture model performance under the conditions of high-resolution and sparse HD-ST data. 

To address these issues, in this paper, we propose a novel image-to-image training and inference paradigm. Unlike traditional spot-wise approaches, our method adopts a region-based sampling strategy, as shown in Figure~\ref{fig:Demo}. Specifically, for each WSI, we select a large spatial window covering multiple bins. The corresponding pathological image patch is used as input to jointly predict gene expression for all bins within the region, thus reducing redundant convolutions and feature copying and achieving near-linear acceleration in both training and inference. Moreover, we further propose Img2ST-Net, which replaces the original regression head that maps each image patch to a single-spot gene expression value with a UNet-based upsampling decoder. This enables the model to perform region-wise prediction of continuous, high-resolution gene expression maps from the entire pathology image in an image-to-image manner. Finally, to overcome the limitations of conventional evaluation metrics in HD-ST data, we propose SSIM-ST, which refers to Spatial Structural Similarity for Spatial Transcriptomics. By computing the structural similarity between predicted and ground-truth expression in spatially aggregated blocks, SSIM-ST mitigates the influence of zero-expression noise and provides a more robust and interpretable performance assessment. Our contribution can be concluded as threefold:

\textbullet\ We propose a novel image-to-image paradigm with region-based sampling, enabling efficient and scalable gene expression prediction across high-resolution spatial windows.

\textbullet\ We introduce Img2ST-Net, a fully convolutional image-to-image
approach that models HD ST data as super-pixel representations and generates region-wise gene expression maps, replacing traditional single-spot regression with dense spatial decoding.

\textbullet\ We design SSIM-ST, a new evaluation metric that measures structural similarity in spatial blocks, offering a more robust and interpretable assessment for HD-ST data.
\section{Related Work}
\subsection{Deep Learning Approaches for ST Prediction}

Spatial transcriptomics (ST) technologies enable high-resolution mapping of gene expression within intact tissue architecture, bridging histological context and molecular profiling~\cite{staahl2016visualization, wang2018three, eng2019transcriptome}. Concurrently, advances in deep learning have revolutionized medical image analysis~\cite{deng2025casc, ke2024tshfna, zhu2025cross,deng2025irs}, leading to a variety of methods for predicting spatial gene expression~\cite{he2020integrating, yang2023exemplar, xie2024spatially, pang2021leveraging, zeng2022spatial}.

ST-Net \cite{he2020integrating} was the first to tackle this task by using DenseNet121~\cite{huang2017densely} in a transfer-learning framework to perform patch-to-spot regression, predicting gene expression for individual ST spots. Following this, similarity-based methods emerged. EGN \cite{yang2023exemplar} employs an exemplar learning mechanism to identify and leverage the most visually similar location within a WSI for each query spot. BLEEP \cite{xie2024spatially} constructs a bi-modal embedding space to match and infer expression profiles from nearest neighbors. While effective, these approaches can be sensitive to staining variability and sample heterogeneity, which may hamper their performance on unseen data. In contrast, spatial-aware approaches explicitly model inter-spot relationships. HisToGene \cite{pang2021leveraging} and His2ST \cite{zeng2022spatial} integrate Transformers and graph neural networks, respectively, to capture both global context and local neighborhood interactions. TRIPLEX \cite{chung2024accurate} extends this concept by incorporating multi-resolution features, combining patch-level and whole-slide information to further boost prediction accuracy. 

In recent years, emerging studies have offered novel perspectives for tackling ST tasks. Diffusion-based STEM~\cite{zhu2025diffusion} reframes spatial gene expression prediction as a conditional denoising diffusion process, iteratively refining noisy expression maps by leveraging contextual features from histopathological images. ASIGN~\cite{zhu2025asign} introduced a new learning paradigm that shifts from 2D WSI-ST predictions to partially known 3D volumetric WSI-ST imputation. It converts isolated 2D spatial relationships into a cohesive 3D structure through cross-layer overlap and similarity, integrating 3D spatial information into ST prediction. In parallel, OmiCLIP~\cite{chen2025visual} builds upon a vision-language foundation model trained with large-scale pretraining, aligning histopathological image features with omics-level representations and leveraging the generalization ability of multimodal pretraining for downstream ST tasks.

\subsection{Predictive Modeling for High-Resolution Visium HD Data}

Recent advancements in ultra-high-resolution ST technologies, such as Visium HD, introduce unique computational and analytical challenges distinct from conventional datasets. To address these complexities, several specialized models have been proposed. MagNet~\cite{zhu2025magnet}, for instance, employs a multi-level attention graph network to leverage hierarchical graph structures across multiple spatial scales. By effectively integrating multi-scale image features, MagNet achieves substantial improvements in both computational efficiency and prediction accuracy at the bin-level compared to traditional approaches.

Similarly, Bin2Cell~\cite{Bin2Cell2024} proposes a cell-centric reconstruction method tailored specifically for high-resolution data. This approach begins by segmenting cell nuclei from histological images and subsequently aggregates neighboring Visium HD bins into coherent cellular units via a graph-based modeling framework. The expression profiles derived from these HD bins are then accurately mapped onto the reconstructed cells, enabling precise, cell-level gene expression profiling and supporting detailed downstream spatial biology analyses.

However, most existing methods follow a one-to-one patch-to-spot prediction paradigm, where each input image patch is independently regressed to the gene expression vector of its corresponding spot. While this design performs well for conventional ST data, it becomes computationally intensive and inefficient when applied to ultra-high-resolution HD datasets. In such scenarios, patch-level prediction leads to substantial redundancy in convolutional operations and feature extraction, as adjacent patches often overlap significantly yet are processed separately.

In contrast, our proposed image-to-image region-wise prediction framework fundamentally departs from this paradigm by adopting a region-based modeling strategy. Instead of treating each spot or bin as an isolated target, the model takes a larger image region as input and jointly predicts gene expression for all bins within that region in a single forward pass. Our proposed method not only captures richer spatial context but also enables near-linear acceleration during both training and inference, significantly enhancing scalability and efficiency for high-resolution HD data.
\section{Method}

\begin{figure*}[htbp]
    \centering
    \includegraphics[width=\linewidth]{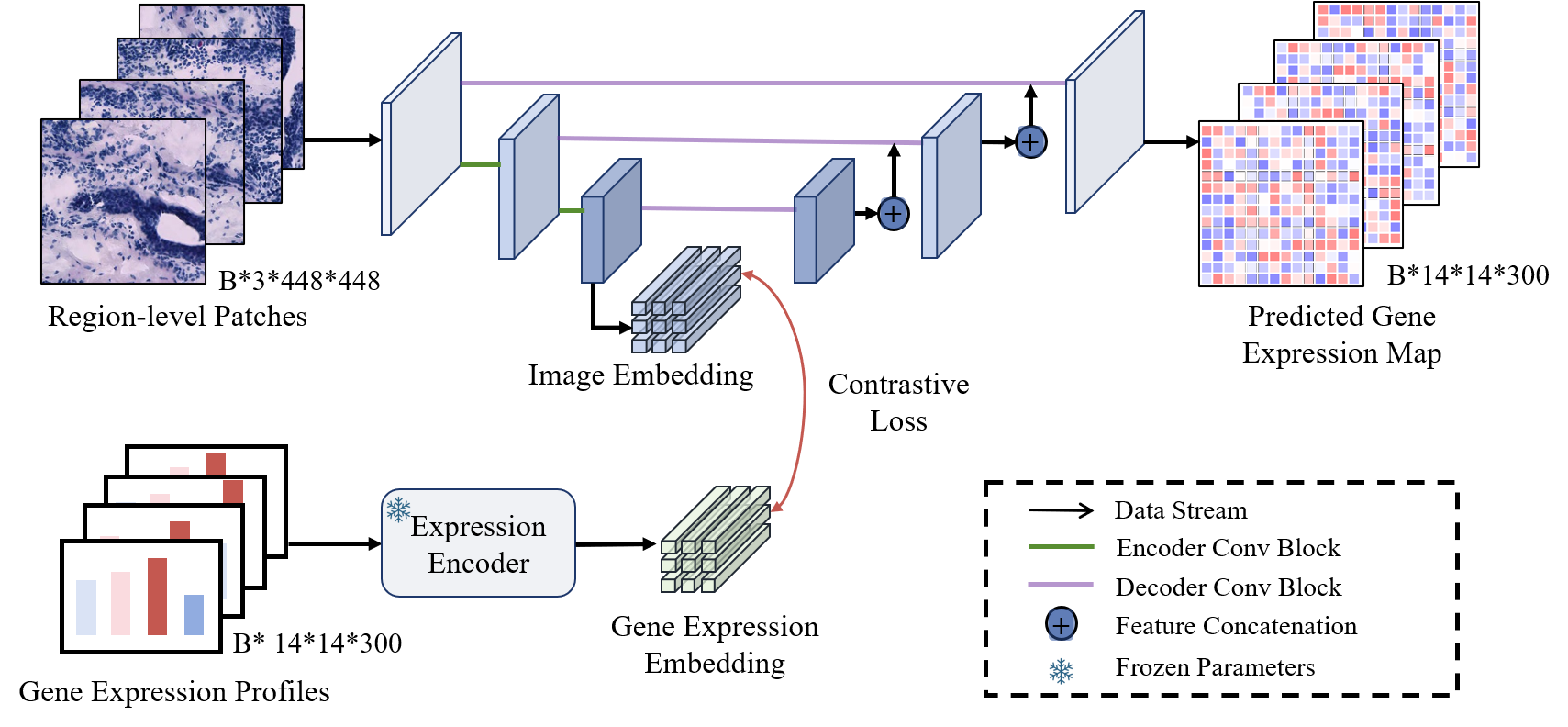}
    \caption{\textbf{Overall framework of our proposed Img2ST-Net.}Fig 2: Overall framework of our proposed Img2ST-Net. Region-level histological patches are processed through a UNet-based architecture to predict high-resolution gene expression maps. Simultaneously, regional gene expression profiles are encoded into embeddings using a frozen expression encoder. A contrastive loss aligns these embeddings with intermediate image features, facilitating accurate and efficient spatial gene expression prediction.}
    \label{fig:framework}
\end{figure*}

\subsection{Fully-convolutional Patch-to-Gene Expression Mapping}

Conventional approaches in ST prediction typically adopt a point-wise modeling paradigm, where each sampled spot is paired with its corresponding image patch to predict gene expression. However, they suffer from two major limitations in practice. First, the inference process is computationally expensive, making point-wise prediction a severe bottleneck in scalability. Second, it fails to exploit the inherent correlation among spots within local neighborhoods. Adjacent positions often share similar histological characteristics, and treating each spot independently neglects valuable regional contextual information.

To address these limitations, we propose a region-level image-to-image prediction strategy. The overall framework of our proposed Img2ST-Net is shown in Figure~\ref{fig:framework} Specifically, we divide each tissue slide into fixed-size regional patches, where each patch is represented as $\mathbf{X} \in \mathbb{R}^{3 \times H \times W}$, with $H = W$ denoting the size of the input patch. The objective of this module is to directly predict the gene expression map across all positions within the region, denoted as $\hat{\mathbf{Y}} \in \mathbb{R}^{C \times H' \times W'}$, where $C$ is the number of target genes and $H' \times W'$ corresponds to the number of the bins within this region.

We adopt a UNet architecture as the backbone for regional prediction. UNet consists of a symmetric encoder-decoder structure that captures features while preserving fine-grained spatial information. The encoder is defined as:

$$
\mathbf{F}_l = f_l^{\text{enc}}(\mathbf{F}_{l-1}), \quad \mathbf{F}_0 = \mathbf{X}, \quad l = 1, 2, \ldots, L
$$

Each $f_l^{\text{enc}}$ comprises two convolutional layers and a downsampling operation to extract hierarchical spatial features. Then, in the decoder, upsampling operations are applied progressively to recover spatial resolution. Skip-connections are used to merge encoder features from corresponding levels:

$$
\mathbf{D}_l = f_l^{\text{dec}}\left(\text{Concat}\left(\text{Up}(\mathbf{D}_{l-1}), \mathbf{F}_{L-l}\right)\right), \quad l = 1, 2, \ldots, L
$$

Finally, an output convolutional layer maps the decoded features to a dense gene expression map: $\hat{\mathbf{Y}} = f^{\text{out}}(\mathbf{D}_L)$. The output $\hat{\mathbf{Y}}$ represents the predicted expression values for all $C$ genes across each bin within the regional patch.

To supervise the learning process, we employ a pixel-wise mean squared error loss over all genes and spatial locations:

$$
L_{\text{reg}} = \frac{1}{C H' W'} \sum_{c=1}^{C} \sum_{i=1}^{H'} \sum_{j=1}^{W'} \left( \hat{Y}_{cij} - Y_{cij} \right)^2
$$

where $Y_{cij}$ denotes the ground truth gene expression value at location $(i,j)$ for gene $c$, and $\hat{Y}_{cij}$ is the predicted value from the model.

\subsection{Regional Contrastive Alignment}
To enhance the alignment between regional image features and gene embeddings, we introduce a contrastive alignment module. It leverages a frozen gene expression encoder to provide a latent reference space for spatial transcriptomics, guiding the image encoder to learn biologically informative representations.

We first utilize a 2-layer MLP as the gene expression encoder $g(\cdot)$ to embed the raw expression vector of each bin $\mathbf{y}_{ij} \in \mathbb{R}^C$ into a latent embedding space to get the gene expression embedding as $\mathbf{z}_{ij}^{\text{exp}} = g(\mathbf{y}_{ij}) \in \mathbb{R}^d$, where $d$ denotes the dimensionality of the expression embedding. The encoder is frozen during training and serves as a fixed reference in the latent space of gene expression.

Meanwhile, the intermediate feature map from the UNet decoder, denoted as $\mathbf{D}*L \in \mathbb{R}^{C \times H' \times W'}$, is projected into the same embedding space via a lightweight linear projection head $f*{\text{proj}}$. For each spatial location $(i, j)$, the projected image embedding is given by:

$$
\hat{\mathbf{z}}_{ij}^{\text{img}} = f_{\text{proj}}(\mathbf{D}_{L,:,i,j}) \in \mathbb{R}^d
$$

We adopt an InfoNCE-based contrastive loss to minimize the distance between each image-predicted embedding $\hat{\mathbf{z}}_{ij}^{\text{img}}$ and its corresponding expression embedding $\mathbf{z}_{ij}^{\text{exp}}$, while pushing away negative pairs. For each positive pair $(i, j)$, the contrastive loss is defined as:

$$
L_{ij}^{\text{contrast}} = - \log \frac{
\exp\left( \text{sim}(\hat{\mathbf{z}}_{ij}^{\text{img}}, \mathbf{z}_{ij}^{\text{exp}}) / \tau \right)
}{\sum\limits_{(i',j')} \exp\left( \text{sim}(\hat{\mathbf{z}}_{ij}^{\text{img}}, \mathbf{z}_{i'j'}^{\text{exp}}) / \tau \right)
}
$$

where $\text{sim}(\cdot, \cdot)$ denotes the cosine similarity function and $\tau$ is a temperature hyperparameter.

The overall contrastive loss across the entire region is computed by averaging over all bins:

$$
L_{\text{contrast}} = \frac{1}{H' W'} \sum_{i=1}^{H'} \sum_{j=1}^{W'} L_{ij}^{\text{contrast}}
$$

To jointly optimize both pixel-level prediction accuracy and semantic alignment, we combine the regression loss with the contrastive loss to form the final training objective:

$$
L_{\text{total}} = L_{\text{reg}} + \lambda \cdot L_{\text{contrast}}
$$

where $\lambda$ is a hyperparameter controlling the strength of the contrastive objective.

\subsection{Structural-Aware Evaluation Metrics}
Pearson Correlation Coefficient (PCC) is commonly used to evaluate the spatial correlation between the predicted gene expression $\hat{\mathbf{y}}$ and the ground truth $\mathbf{y}$. It is defined as:

$$
\text{PCC}(\mathbf{y}, \hat{\mathbf{y}}) = \frac{\sum_{k=1}^{N} (y_k - \bar{y})(\hat{y}_k - \bar{\hat{y}})}{\sqrt{\sum_{k=1}^{N} (y_k - \bar{y})^2} \cdot \sqrt{\sum_{k=1}^{N} (\hat{y}_k - \bar{\hat{y}})^2}}.
$$

However, in high-resolution spatial transcriptomics settings, each WSI typically contains more than $10^6$ bins, and most gene expression values are zero. This leads to an overall mean $\bar{y} \approx 0$ and variance $\mathrm{Var}(y) \approx 0$, making the denominator in the PCC equation approach zero and resulting in numerical instability. When the model introduces small noise $\varepsilon_k \sim \mathcal{N}(0, \sigma^2)$ in regions where the true expression is zero, the correlation can be dominated by noise, even if predictions in key expression regions are accurate. The product terms $(y_k - \bar{y})(\hat{y}_k - \bar{\hat{y}})$ are either zero or dominated by noise, leading to:

$$
\text{PCC}(\mathbf{y}, \hat{\mathbf{y}}) \approx 0,
$$

To address this issue, we introduce the Structural Similarity Index (SSIM) as a complementary metric. SSIM is defined as:

$$
\text{SSIM}(x, y) = \frac{(2\mu_x \mu_y + C_1)(2\sigma_{xy} + C_2)}{(\mu_x^2 + \mu_y^2 + C_1)(\sigma_x^2 + \sigma_y^2 + C_2)},
$$

where $\mu_x$ and $\mu_y$ are local means, $\sigma_x^2$ and $\sigma_y^2$ are variances, and $\sigma_{xy}$ is the covariance between the two local patches. $C_1$ and $C_2$ are small constants to stabilize the division. Unlike PCC, which captures global trends, SSIM operates on local windows and is more sensitive to spatial patterns such as edges and structures. It better reflects the spatial coherence of region-specific expression. As long as the shape and location of expression hotspots are consistent with the ground truth, SSIM can provide stable and reliable assessments.
\section{Data and Experiment}
\noindent\textbf{Dataset.}
We evaluated the proposed Img2ST-Net model and several representative baseline methods on two publicly available spatial transcriptomics datasets: the Breast Cancer dataset~\cite{nagendran2023visium} and the Colorectal Cancer (CRC) dataset~\cite{oliveira2025high}. Both datasets provide spatial transcriptomic profiles at three resolutions: 2~$\mu$m, 8~$\mu$m, and 16~$\mu$m. Due to the extremely sparse gene expression at the 2~$\mu$m resolution, we conducted experiments only at the 8~$\mu$m and 16~$\mu$m resolutions. The Breast Cancer dataset consists of four single-section samples, while the CRC dataset contains five single-section samples, including three CRC tumor tissues and two adjacent normal tissues.

\noindent\textbf{Data Preprocessing.}
We designed two different preprocessing pipelines corresponding to two experimental settings: the traditional one-to-one prediction task and our proposed image-to-image prediction framework. For the one-to-one setting, 112$\times$112 pixel image patches were cropped from bins at the 8~$\mu$m and 16~$\mu$m resolutions and paired with their corresponding gene expression values as model inputs. For the image-to-image setting, we extracted 448$\times$448 pixel patches from WSIs and paired them with all spatial spots (bins) located within the patch area, based on the distance between each bin center and the patch center. At 8~$\mu$m resolution, each patch covers 196 bins; at 16~$\mu$m resolution, each patch covers 49 bins. Following the strategy used in ST-Net~\cite{he2020integrating}, we selected the top 250 genes with the highest average expression from over 20,000 original genes as prediction targets. Gene expression values were normalized using the method adopted in BLEEP~\cite{xie2024spatially}, applying a log transformation to the raw count values to mitigate expression imbalance.

\noindent\textbf{Compared Methods and Evaluation Metrics.}
We conducted a comprehensive comparison between our Img2ST-Net and several state-of-the-art spatial transcriptomics methods, including the spatially aware models HisToGene~\cite{pang2021leveraging} and His2ST~\cite{zeng2022spatial}, the similarity-based method BLEEP~\cite{xie2024spatially}, and the classical baseline ST-Net~\cite{he2020integrating}. All baseline models were implemented using their publicly released official code and evaluated under the same experimental settings.
 To assess performance from both numerical accuracy and spatial consistency perspectives, we employed four evaluation metrics: mean squared error (MSE), mean absolute error (MAE), and structural similarity index (SSIM).

\noindent\textbf{Experimental Setup and Implementation Details.}
All experiments were conducted on 4 NVIDIA RTX A6000 GPUs. We used stochastic gradient descent (SGD) as the optimizer, with a momentum of 0.9 and a weight decay of $10^{-4}$. The initial learning rate was set to $10^{-4}$ and decayed following a cosine annealing schedule down to 1\% of the initial value. All models were trained until convergence. The training batch size was set to 64, and the hyperparameter $\lambda$ in the hybrid loss function was tuned and fixed at 0.25. For evaluation, one sample was randomly selected as the test set in an 8:2 train-test split to assess the generalization performance.
\section{Results}
\subsection{Experimental Results}

\begin{table*}[tbph]
\centering
\caption{\textbf{Quantitative comparisons across different datasets and resolutions.} 
The best performance is highlighted in \textbf{bold}, where we can observe that \textit{Img2ST-Net} outperforms state-of-the-art methods on both Breast Cancer and CRC datasets across most evaluation metrics.}
\begin{tblr}{
  row{1} = {c},
  row{2} = {c},
  cell{1}{1} = {r=2}{},
  cell{1}{2} = {r=2}{},
  cell{1}{3} = {c=3}{},
  cell{1}{6} = {c=3}{},
  cell{3}{1} = {r=5}{c},
  cell{3}{3} = {c},
  cell{3}{4} = {c},
  cell{3}{5} = {c},
  cell{3}{6} = {c},
  cell{3}{7} = {c},
  cell{3}{8} = {c},
  cell{4}{3} = {c},
  cell{4}{4} = {c},
  cell{4}{5} = {c},
  cell{4}{6} = {c},
  cell{4}{7} = {c},
  cell{4}{8} = {c},
  cell{5}{3} = {c},
  cell{5}{4} = {c},
  cell{5}{5} = {c},
  cell{5}{6} = {c},
  cell{5}{7} = {c},
  cell{5}{8} = {c},
  cell{6}{3} = {c},
  cell{6}{4} = {c},
  cell{6}{5} = {c},
  cell{6}{6} = {c},
  cell{6}{7} = {c},
  cell{6}{8} = {c},
  cell{7}{3} = {c},
  cell{7}{4} = {c},
  cell{7}{5} = {c},
  cell{7}{6} = {c},
  cell{7}{7} = {c},
  cell{7}{8} = {c},
  cell{8}{1} = {r=5}{c},
  cell{8}{3} = {c},
  cell{8}{4} = {c},
  cell{8}{5} = {c},
  cell{8}{6} = {c},
  cell{8}{7} = {c},
  cell{8}{8} = {c},
  cell{9}{3} = {c},
  cell{9}{4} = {c},
  cell{9}{5} = {c},
  cell{9}{6} = {c},
  cell{9}{7} = {c},
  cell{9}{8} = {c},
  cell{10}{3} = {c},
  cell{10}{4} = {c},
  cell{10}{5} = {c},
  cell{10}{6} = {c},
  cell{10}{7} = {c},
  cell{10}{8} = {c},
  cell{11}{3} = {c},
  cell{11}{4} = {c},
  cell{11}{5} = {c},
  cell{11}{6} = {c},
  cell{11}{7} = {c},
  cell{11}{8} = {c},
  cell{12}{3} = {c},
  cell{12}{4} = {c},
  cell{12}{5} = {c},
  cell{12}{6} = {c},
  cell{12}{7} = {c},
  cell{12}{8} = {c},
  vline{2-3} = {3-7,8-12}{},
  vline{3} = {4-7,9-12}{},
  hline{1,3,8,13} = {-}{},
  hline{2} = {3-8}{},
}
Resolution & Model           & Breast Cancer~\cite{nagendran2023visium} &        &        & CRC~\cite{oliveira2025high}    &        &        \\
           &                 & MSE            & MAE    & SSIM-ST    & MSE    & MAE    & SSIM-ST   \\
8~$\mu$m & ST-Net~\cite{he2020integrating}          & 0.0990         & 0.2265 & 0.0141 & 0.2858 & 0.3235 & 0.0012 \\
    & His2ST~\cite{zeng2022spatial}          & 0.3461         & 0.4670 & 0.0219 & 0.546  & 0.5461 & 0.0091 \\
     & HisToGene~\cite{pang2021leveraging}       & 0.2917         & 0.4277 & 0.0302 & 0.4809 & 0.5079 & 0.0119 \\
    & BLEEP~\cite{xie2024spatially}            & 0.0823         & 0.1822 & \textbf{0.1285} & 0.2750 & \textbf{0.2218 }& 0.0139 \\
    & \textbf{Img2ST-Net (Ours)} & \textbf{0.0647  }       & \textbf{0.1737} & 0.0756 & \textbf{0.2587} & 0.2711 & \textbf{0.0144} \\
16~$\mu$m& ST-Net~\cite{he2020integrating}          & 0.2059         & 0.3233 & 0.0045 & \textbf{0.7690} & 0.5437 & 0.0038 \\
    & His2ST~\cite{zeng2022spatial}          & 0.4253         & 0.5035 & 0.0206 & 1.0467 & 0.7263 & 0.0038 \\
    & HisToGene~\cite{pang2021leveraging}       & 0.3046         & 0.4260 & 0.0215 & 0.8469 & 0.6383 & 0.0076 \\
     & BLEEP~\cite{xie2024spatially}  & 0.3933         & 0.4671 & 0.0266 & 0.8156 & 0.5227 & \textbf{0.0492} \\
    & \textbf{Img2ST-Net (Ours)} & \textbf{0.1657 }       & \textbf{0.2506} & \textbf{0.0937 }& 0.7981 & \textbf{0.5208} & 0.0081 
\end{tblr}
\label{tab:comparison}
\end{table*}

We conducted evaluations on two public Visium HD datasets under 8~$\mu$m and 16~$\mu$m resolution settings to benchmark Img2ST-Net against state-of-the-art methods. 
Quantitative comparisons across different datasets and resolutions are summarized in Table~\ref{tab:comparison}. At 8~$\mu$m, Img2ST-Net achieves the lowest MSE of 0.0647 on the Breast Cancer dataset and the highest SSIM of 0.0144 on the CRC dataset, reflecting its ability to generate accurate predictions while preserving spatial expression patterns. At 16~$\mu$m, where competing models experience notable performance degradation, Img2ST-Net maintains reliable accuracy. For breast cancer, its SSIM reaches 0.0937, significantly outperforming other methods. These results demonstrate that Img2ST-Net not only delivers superior predictive performance but also exhibits strong robustness to resolution changes, with enhanced capacity for spatial structure preservation.

\begin{figure*}[htbp]
    \centering
    \caption{\textbf{Mean expression comparison across models and resolutions.} 
We visualize the average expression of each gene, sorted by ground truth values at two resolutions for the CRC~\cite{oliveira2025high} dataset. 
Each plot compares the predicted mean expression (orange dots) against the ground truth (blue curve). 
\textit{Img2ST-Net} achieves the closest alignment with the ground truth distribution, demonstrating better modeling of global gene expression trends across both resolutions.}
    \includegraphics[width=\linewidth]{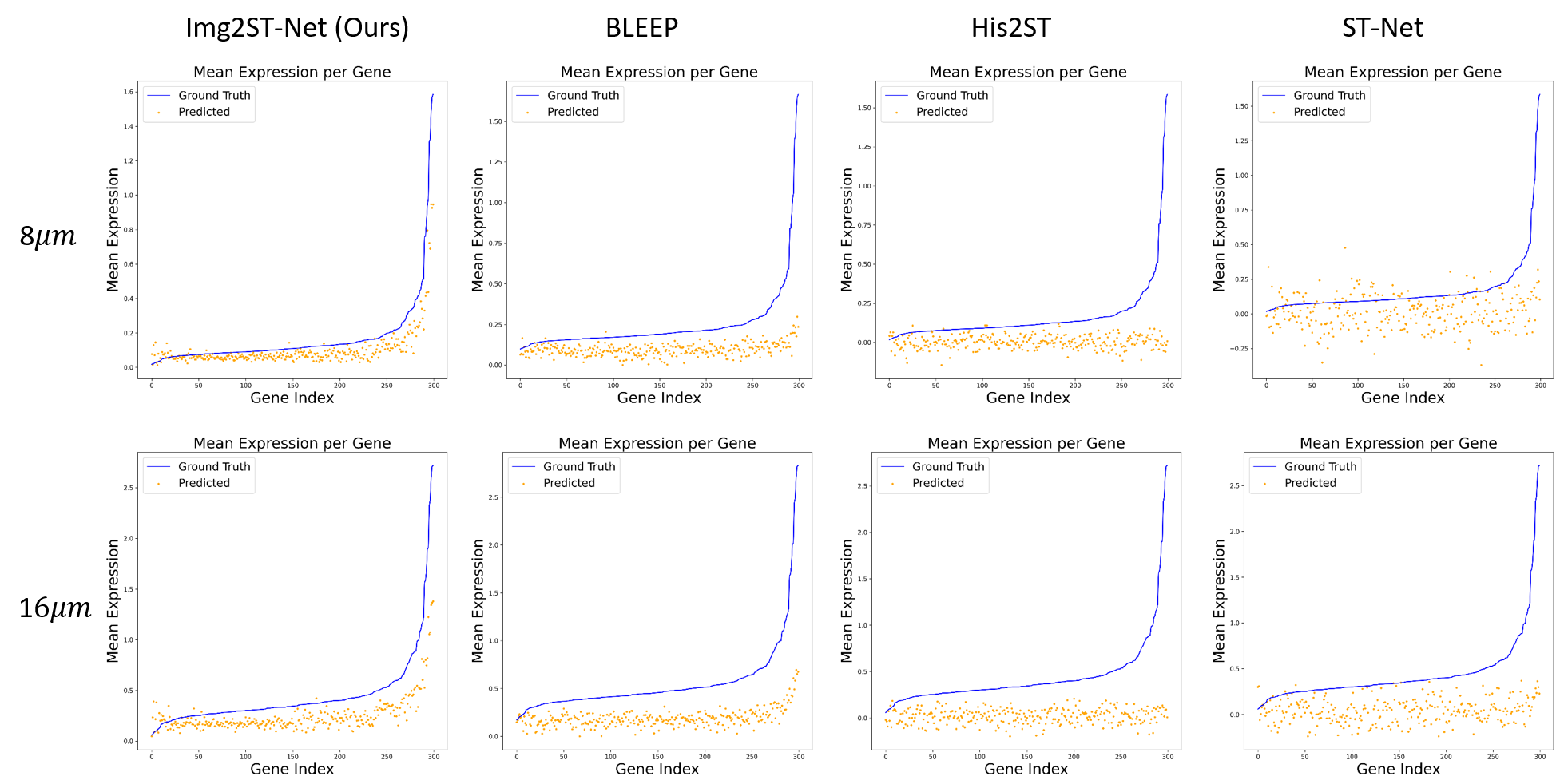}
    \label{fig:gene-dist}
\end{figure*}

To further examine the alignment of predicted gene expression with ground truth, we visualize the sorted per-gene mean expression distributions in Figure~\ref{fig:gene-dist}. Each subplot compares the predicted mean expression (orange dots) to the ground truth distribution (blue curve), sorted by ground truth magnitude. Across both 8~$\mu$m and 16~$\mu$m resolutions, Img2ST-Net demonstrates the closest adherence to the true distribution, capturing both low- and high-expression gene patterns effectively. In contrast, baseline methods show noticeable deviation or compression in the predicted expression ranges, especially at lower resolutions. This alignment with ground truth distribution corroborates the quantitative results and further validates the expressiveness and stability of Img2ST-Net.

\begin{table*}[tbph]
\centering
\caption{\textbf{Ablation study for functional blocks in Img2ST-Net on Breast Cancer dataset~\cite{nagendran2023visium}.} 
Our proposed \textit{Image-to-image Setting} significantly reduces training time while preserving prediction accuracy. Img2ST-Net achieves optimal results when integrating all modules.}
\begin{tblr}{
  row{1} = {c},
  cell{2}{1} = {r=3}{c},
  cell{2}{3} = {c},
  cell{2}{4} = {c},
  cell{2}{5} = {c},
  cell{2}{6} = {c},
  cell{3}{3} = {c},
  cell{3}{4} = {c},
  cell{3}{5} = {c},
  cell{3}{6} = {c},
  cell{4}{3} = {c},
  cell{4}{4} = {c},
  cell{4}{5} = {c},
  cell{4}{6} = {c},
  cell{5}{1} = {r=3}{c},
  cell{5}{3} = {c},
  cell{5}{4} = {c},
  cell{5}{5} = {c},
  cell{5}{6} = {c},
  cell{6}{3} = {c},
  cell{6}{4} = {c},
  cell{6}{5} = {c},
  cell{6}{6} = {c},
  cell{7}{3} = {c},
  cell{7}{4} = {c},
  cell{7}{5} = {c},
  cell{7}{6} = {c},
  vline{2-3} = {2-4,5-7}{},
  vline{3} = {2-4,5-7}{},
  hline{1-2,5,8} = {-}{},
}
Resolution & Functional Blocks~~    & Runtime (min/epoch) & MSE    & MAE    & SSIM-ST    \\
8~$\mu$m & w.o. Contrastive Loss  & /                   & 0.1145 & 0.2404 & \textbf{0.0766 }\\
           & One-to-One Setting~    & 12.10               & 0.0661 & 0.1757 & 0.0142 \\
           & \textbf{Image-to-image Setting} &\textbf{ 0.43  }              & \textbf{0.0647} & \textbf{0.1737} & 0.0756 \\
16~$\mu$m  & w.o. Contrastive Loss  & /                   & 0.2118 & 0.3512 & 0.0405 \\
           & One-to-One Setting~    & 3.25                  & \textbf{0.1655} & \textbf{0.2504} & 0.0044 \\
           & \textbf{Image-to-image Setting} & \textbf{0.43  }              & 0.1657 & 0.2506 & \textbf{0.0937 }
\end{tblr}
\label{tab:ablation}
\end{table*}

\subsection{Ablation Study}
To understand the contribution of each component in Img2ST-Net, we conduct an ablation study on the Breast Cancer dataset at both 8~$\mu$m and 16~$\mu$m resolutions. The results are summarized in Table~\ref{tab:ablation}. Removing the contrastive loss leads to a clear drop in performance across all metrics, especially in SSIM, which indicates reduced spatial coherence in the predicted expression maps. At 16~$\mu$m resolution, removing the contrastive loss increases the MSE from  0.1657 to 0.2118 and decreases SSIM from 0.0937 to 0.0405, showing its importance in preserving spatial consistency.

We also compare the proposed \textit{Image-to-image Setting} with the conventional one-to-one formulation. While both settings achieve comparable accuracy, our image-to-image strategy drastically improves training efficiency. At 8~$\mu$m, the image-to-image setting reduces training time per epoch from 12.10 minutes to just 0.43 minutes, resulting in approximately \textbf{28-fold acceleration}. The combination of these components enables Img2ST-Net to achieve superior performance with significantly improved training efficiency.

\section{Conclusion}
We introduce a comprehensive framework that rethinks spatial transcriptomics modeling for the high-resolution era. Moving beyond the limitations of conventional spot-wise prediction, our region-based image-to-image paradigm efficiently captures gene expression patterns over dense spatial grids, offering substantial gains in both scalability and spatial fidelity. We propose Img2ST-Net, a UNet-style architecture that transforms pathological image regions into high-resolution gene expression maps, enabling context-aware and fine-grained modeling across large tissue areas. To complement this paradigm shift, we propose SSIM-ST, a novel evaluation metric grounded in structural similarity, which robustly quantifies prediction quality even under severe sparsity and noise. Collectively, our contributions provide a scalable modeling solution and a principled evaluation strategy, laying the groundwork for accurate, efficient, and biologically coherent analysis in next-generation spatial transcriptomics.

\section*{Code, Data, and Materials Availability}
The code is available on GitHub. All datasets are publicly available.

\section* {Acknowledgments}
This research was supported by NIH R01DK135597 (Huo), DoD HT9425-23-1-0003 (HCY), and KPMP Glue Grant. This work was also supported by Vanderbilt Seed Success Grant, Vanderbilt Discovery Grant, and VISE Seed Grant. This project was supported by The Leona M. and Harry B. Helmsley Charitable Trust grant G-1903-03793 and G-2103-05128. This research was also supported by NIH grants R01EB033385, R01DK132338, REB017230, R01MH125931, and NSF 2040462. We extend gratitude to NVIDIA for their support by means of the NVIDIA hardware grant. This work was also supported by NSF NAIRR Pilot Award NAIRR240055.

\bibliography{report} 
\bibliographystyle{spiebib} 

\end{document}